\documentclass[runningheads]{llncs}

\usepackage{graphicx}
\usepackage{todonotes}
\usepackage{amsmath}
\usepackage{amssymb}
\usepackage{algorithmic}
\usepackage{array}
\usepackage{eqparbox}
\usepackage{url}
\usepackage{algorithm2e}
\SetKw{Add}{Add}
\SetKw{Return}{Return}
\usepackage{multirow}
\usepackage{booktabs}
\usepackage{tabularx}
\usepackage{comment}
\usepackage{capt-of}
\usepackage{multirow}
\usepackage[caption=false]{subfig}
\usepackage{float}
\usepackage{tikz}
\usetikzlibrary{shapes.geometric, arrows.meta, positioning}

\newcounter{equationset}
\begin{document}

\title{\LARGE Pack and Extract: An Effective Graph-Based Approach for Feature Selection}
\titlerunning{Graph-Based Approach for Feature Selection}

%\author{Faisal N. Abu-Khzam\inst1
\author{Faisal N. Abu-Khzam\inst1\thanks{Corresponding author: faisal.abukhzam@lau.edu.lb}
\and Joseph R. Barr\inst2
\and Cynthia Chaaya\inst1
\and Nadim Obeid\inst1
\and Peter Shaw\inst3}
\authorrunning{Abu-Khzam et al.}
\institute{
Lebanese American University, 
Beirut, Lebanon\\
\and
Embry-Riddle Aeronautical University, 
Prescott, Arizona, USA\\
\and 
Oujiang Laboratory, Wenzhou, Zhejiang, China
}

\maketitle 

%\keywords{Feature selection, $d$-Packing, Weighted Independent Dominating Set, Classification, and Diversity.}

\begin{abstract}
Many molecular biology data sets contain thousands of features, such as genes in single-cell experiments or other omics data. Standard predictive models often perform poorly in such settings, as they tend to retain a large number of features and fail to adequately handle correlations among them. Combinatorial approaches, such as feature selection based on the \textsc{Independent Dominating Set} (IDS) model, have shown promising results, but appear to have been largely overlooked, possibly due to their high computational complexity.

In this paper, we propose an improvement to IDS-based feature selection that (i) favors features exhibiting greater diversity and (ii) generalizes the notion of independent dominating set to a node packing 
model. The resulting formulation is used as an intermediate step within the feature selection process. We evaluate our approach on five genetic data sets and compare the results with those obtained using existing methods. Our experimental results demonstrate notable improvements in predictive accuracy over traditional models, as well as increased stability of the selected feature sets.
\end{abstract}

\thispagestyle{empty}

\section{Introduction}

The principle of parsimony, or Occam’s razor, is fundamental
%Ockham's razor, or the principle of parsimony, is fundamental 
in philosophy and the natural sciences, and plays a central role in modeling physical phenomena and, more generally, in statistical modeling. In statistical learning, this principle manifests itself most clearly through {\em feature selection}, whose goal is to identify a small subset of informative variables while discarding redundancy and noise.

Feature selection seeks to preserve or improve predictive performance while reducing overfitting, enhancing interpretability, and lowering computational complexity. Existing approaches can be broadly classified as supervised or unsupervised. Supervised methods include techniques based on statistical tests such as the F-test and the chi-squared test, as well as information-theoretic and model selection criteria such as the Akaike and Bayesian Information Criteria. Unsupervised methods include dimensionality reduction and redundancy control techniques such as principal component analysis and variance inflation factor (VIF)-based approaches.

A wide range of heuristic strategies has also been proposed. Some rely on purity criteria, such as minimizing the Gini coefficient, and are commonly used in decision tree learning. Ensemble methods, including random forests and boosting, combine multiple decision trees to improve predictive performance. Other learning frameworks, such as {\em support vector machines (SVMs)}, rely on constrained optimization formulations that minimize convex loss functions subject to margin constraints. More recently, {\em regularization methods}, such as the {\em lasso} procedure, have become popular for shrinking the effective input dimension and controlling model complexity \cite{Lasso}.

In the simplest setting involving two variables, pairwise inspection using Pearson’s correlation coefficient may suffice to detect dependency, and removing one of two highly correlated variables is often adequate. However, this approach fails to capture correlations between a variable and a group of other variables. The variance inflation procedure partially addresses this limitation by using the {\em variance inflation factor (VIF)}, which relies on the coefficient of determination ($R^2$) from multiple linear regression to identify variables that are correlated with subsets of other variables. Nevertheless, such approaches remain limited in their ability to capture more global dependency structures.

An alternative perspective consists of modeling feature dependencies using graphs. In this setting, features correspond to vertices, and edges encode pairwise correlations exceeding a prescribed threshold. Graph-based representations naturally capture structural relationships among features and enable the use of combinatorial optimization techniques for feature selection. In particular, selecting a minimum subset of features that dominates the correlation graph while remaining mutually independent leads to the {\em Independent Dominating Set} (\textsc{IDS}) model, which was proposed more than a decade ago \cite{shu2013}. Despite promising empirical results, this approach has received limited attention, possibly due to the reliance on integer linear programming formulations with exponential worst-case complexity.

Feature selection is especially critical in medical and biological applications, where data sets often exhibit extremely high dimensionality and strong correlations among variables. Reducing dimensionality in such settings facilitates both predictive modeling and the interpretation of underlying biological mechanisms \cite{kawamoto2005improving,awan2019feature}. Empirical studies have also shown that no single feature selection method consistently outperforms others across all evaluation criteria, highlighting the importance of combining complementary perspectives and performance measures \cite{KOU2020105836}.

Motivated by these observations, we revisit graph-based feature selection and propose a more flexible combinatorial framework that extends the \textsc{IDS} model. Rather than restricting attention to immediate adjacency, we consider a generalized approach based on the notion of a \emph{node packing}, or $d$-packing \cite{abs-2401-00525}, which enforces separation between selected features at a prescribed graph distance. To differentiate between correlated features, we further incorporate a diversity criterion that favors informative variables, allowing the model to select representative features from correlated groups.

\paragraph{Contributions.}
The contributions of this work are fourfold. First, we introduce a diversity-driven weighting scheme based on mutual information to guide the selection of informative features within correlated groups. Second, we formulate graph-based feature selection as a weighted $d$-packing problem on correlation graphs, generalizing independent dominating set–based models. Third, we develop an efficient greedy algorithm inspired by maximum-utility heuristics, thereby avoiding reliance on integer linear programming formulations. Finally, we demonstrate on multiple high-dimensional biological data sets that the proposed graph-based preselection, when combined with standard shrinkage methods, improves both predictive accuracy and the stability of selected feature sets.

\section{Preliminaries}

We assume familiarity with basic graph-theoretic terminology. A {\em dominating set} in a graph $G=(V,E)$ is a set $D \subseteq V$ such that every vertex in $V \setminus D$ has at least one neighbor in $D$. 
A {\em $d$-packing} in $G$ is a subset $P \subseteq V$ satisfying the following conditions:  
(i) for any $v,w \in P$, the length of a shortest path between $v$ and $w$ is strictly greater than $d$, and 
(ii) for every vertex $v \in V \setminus P$, there exists a vertex in $P$ at distance at most $d$ from $v$.  
The special case $d=1$ corresponds to the notion of a \emph{maximal independent set}. In this context, a set of vertices is independent if no two of its elements are adjacent. The definition of a $1$-packing ensures maximality in the following sense: if $P$ is a $1$-packing in $G$, then (i) no two vertices in $P$ are adjacent, and (ii) no additional vertex can be added to $P$ without violating independence. Hence, $P$ is inclusion-wise maximal (i.e., it is not properly contained in a larger independent set).

The problem of finding a minimum-size maximal independent set is equivalent to the {\sc Independent Dominating Set} (\textsc{IDS}) problem, since every vertex outside the set must be adjacent to a vertex inside it. The \textsc{IDS} problem is a well-studied combinatorial optimization problem \cite{wang2020two} and is known to be $\mathcal{NP}$-hard \cite{karp2010reducibility} and $\mathcal{PLS}$-hard \cite{borzechowski2016complexity}. The asymptotically fastest known exact algorithm for \textsc{IDS} runs in time $\mathcal{O}(1.3351^n)$ \cite{BOURGEOIS2013558}.

\subsubsection*{Feature Selection}
The primary goal of feature selection is to identify a subset of independent variables that captures as much relevant information as possible using as few features as possible. More precisely, the aim is to select a subset of features for which a predictive model is optimal, where optimality is typically assessed using cross-validation.

The term {\em dimensionality reduction} is often used in this context, as training instances $(x,y)$ with $x \in \mathbb{R}^p$ are replaced by $(x',y)$ with $x' \in \mathbb{R}^q$, where $q < p$. An optimal feature set typically leads to a model that achieves a favorable bias–variance trade-off \cite{CHANDRASHEKAR201416,BarrZacks}.
This issue is particularly evident in standardized gene expression data, which often contain a large number of variables with substantial pairwise correlations. In such cases, two highly correlated features can often be replaced by a single representative feature without degrading model performance \cite{CHANDRASHEKAR201416}. Variables that appear irrelevant are frequently dominated by noise, and their removal can improve generalization. Identifying an optimal feature subset becomes especially challenging in high-dimensional settings where the number of features exceeds the number of samples \cite{KHAIRE20221060}.

Let $G=(V,E)$ be a graph constructed from tabular data, where vertices correspond to features and an edge connects two vertices if the corresponding features are deemed ``close'' according to a correlation measure, to be specified later. Computing a minimum $d$-packing in $G$ yields a set of features that are sufficiently uncorrelated, while ensuring that every remaining feature lies within distance at most $d$ from one of the selected features.

In this work, we further assume that features are associated with weights reflecting their statistical significance. Our objective is therefore to identify a $d$-packing of maximum total weight, where the weight of a packing is the sum of the weights of its vertices. The resulting {\sc Minimum Weighted Independent Dominating Set with diameter $d$} problem (also referred to as the \textsc{(w,$d$)-packing} problem) is $\mathcal{NP}$-hard via a straightforward reduction from the unweighted $d$-packing problem.

We assign feature weights based on a notion of {\em diversity}: features exhibiting greater variability with respect to the class label are considered more informative. This concept is discussed in \cite{SHAHEEN2023329}. Since our formulation is minimization-based, lower weights are assigned to more diverse features, or equivalently, the greedy heuristic is adapted to prioritize such features. Further details are provided in the next section.

\subsubsection*{The Role of VC Dimension}
The Vapnik–Chervonenkis (VC) dimension measures the capacity of a hypothesis class, indicating the largest number of points that can be shattered by hypotheses from that class. A higher VC dimension corresponds to increased flexibility and lower bias, but also a greater risk of overfitting when the training sample size is limited.

Within the framework of Probably Approximately Correct (PAC) learning, one can derive bounds on the generalization error. For any hypothesis $h$, with probability at least $1-\delta$, the following holds:
\begin{equation*}
|R(h)-\hat{R}(h)| \leq \sqrt{\frac{2}{m}\left(d \ln\left(\frac{2m}{d}\right)+\ln\left(\frac{4}{\delta}\right)\right)},
\end{equation*}
where $R(h)$ denotes the true risk, $\hat{R}(h)$ the empirical risk on a sample of size $m$, and $d$ the effective VC dimension. Moreover, to achieve an error margin $\epsilon$ with high probability, the sample size must satisfy
\begin{equation*}
m = \Omega\left(\frac{d+\ln(1/\delta)}{\epsilon^2}\right).
\end{equation*}
These bounds highlight the trade-off between model complexity and the amount of data required for reliable learning.

\subsubsection*{Evaluation Metrics}
If a model is too simple, for example by using too few features, it may suffer from high bias and fail to capture important structure in the data. Conversely, overly complex models may exhibit high variance and overfit noise, resulting in poor generalization.

To assess this trade-off, we use several evaluation metrics in addition to accuracy. Specifically, we report:
\begin{itemize}
    \item \textbf{Recall:}
    \[
    \text{Recall} = \frac{\mathrm{TP}}{\mathrm{TP} + \mathrm{FN}}
    \]
    \item \textbf{Precision:}
    \[
    \text{Precision} = \frac{\mathrm{TP}}{\mathrm{TP} + \mathrm{FP}}
    \]
    \item \textbf{F1 score:}
    \[
    \text{F1} = 2 \cdot \frac{\text{Precision} \times \text{Recall}}{\text{Precision} + \text{Recall}}
    \]
\end{itemize}

Barr et al.~\cite{barr2023variability} proposed balancing bias and variance by gradually increasing model complexity until predictive performance peaks, an approach that they formally justified for generalized linear models and supported empirically. 
%This model was later enhanced via the introduction of feature-weights \cite{abu-wids}, which will be adopted in this paper.

In our setting, the $d$-packing parameter $d$ plays a role analogous to a capacity parameter. Smaller values of $d$ impose stricter separation between selected features, leading to higher bias, while larger values allow greater flexibility at the expense of increased variance. Selecting an appropriate value of $d$ thus serves as a form of regularization and can improve overall model performance.

\section{Methodology}

We propose a graph-based feature selection framework that combines combinatorial optimization with statistical learning techniques. The overall approach follows a five-stage pipeline, illustrated in Fig.~\ref{fig:pipeline}. First, the input data set is transformed into a correlation graph. Second, feature-level weights reflecting statistical diversity are assigned to graph vertices. Third, a weighted $d$-packing is computed to obtain a reduced and weakly correlated feature set. Fourth, additional shrinkage-based feature selection methods are applied as a refinement step. Finally, classification models are trained using the selected features.

\tikzstyle{process} = [rectangle, minimum width=3cm, minimum height=1cm, text centered, draw=black, fill=blue!15]
\tikzstyle{arrow} = [thick,->,>=Stealth]

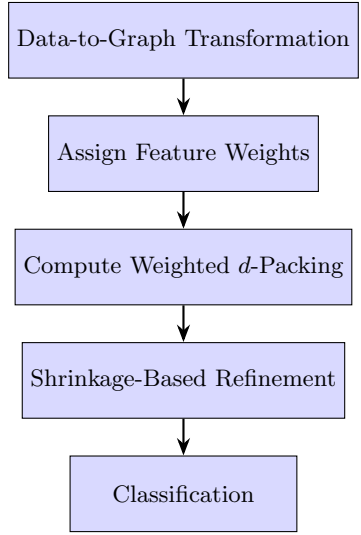
\begin{figure}[H]
    \centering
    \caption{Overview of the proposed feature selection pipeline}
    \label{fig:pipeline}
    \begin{center}
    \begin{tikzpicture}[node distance=1.5cm]
    
    \node (step1) [process] {Data-to-Graph Transformation};
    \node (step2) [process, below of=step1] {Assign Feature Weights};
    \node (step3) [process, below of=step2] {Compute Weighted $d$-Packing};
    \node (step4) [process, below of=step3] {Shrinkage-Based Refinement};
    \node (step5) [process, below of=step4] {Classification};
    
    \draw [arrow] (step1) -- (step2);
    \draw [arrow] (step2) -- (step3);
    \draw [arrow] (step3) -- (step4);
    \draw [arrow] (step4) -- (step5);
    
    \end{tikzpicture}
    \end{center}
\end{figure}

\subsection{Graph Construction}

Following \cite{shu2013}, features are represented as vertices in an undirected graph. Let $V$ denote the set of vertices corresponding to the columns of the data matrix. Two vertices $u,v \in V$ are connected by an edge if the corresponding features exhibit sufficient correlation, defined as
\[
\big|\mathrm{cor}(u,v)\big| \geq \tau,
\]
where $\mathrm{cor}(u,v)$ denotes the Pearson correlation coefficient and $\tau \in (0,1]$ is a user-defined threshold parameter. Pearson correlation is used throughout this work due to its simplicity, interpretability, and widespread use in biological data analysis. The resulting graph captures pairwise dependencies between features and serves as the structural basis for subsequent combinatorial selection.

\subsection{Feature Diversity and Weight Assignment}

To differentiate between correlated features, we associate each vertex with a weight that reflects the feature’s statistical relevance. As a measure of feature diversity, we use mutual information \cite{Info}, which quantifies the dependency between a feature $X$ and the target variable $Y$ and is well suited for continuous-valued data:
\[
I(X;Y) = \sum_{x \in \mathcal{X}} \sum_{y \in \mathcal{Y}} p(x,y)\log\left(\frac{p(x,y)}{p(x)p(y)}\right)
\]
%\equationset{Mutual Information}
\label{eq:MI}

Features exhibiting higher mutual information with respect to the target variable are considered more informative. Since the underlying combinatorial problem is formulated as a minimization problem, lower weights are assigned to more diverse features, thereby biasing the selection process toward informative variables.

\subsection{Weighted $d$-Packing Formulation}

Given the weighted correlation graph $G=(V,E)$ and an integer $d \geq 1$, our objective is to compute a subset $P \subseteq V$ that forms a $d$-packing and optimizes the total weight of the selected vertices. Again, the case $d=1$ corresponds to a weighted independent dominating set, while larger values of $d$ enforce stronger separation between selected features.
Obviously, this formulation generalizes the classical Independent Dominating Set problem, which is $\mathcal{NP}$-hard even in the unweighted case. Consequently, exact methods based on integer linear programming do not scale to high-dimensional feature spaces, motivating the use of efficient heuristic approaches.

\subsection{Greedy Algorithm for Weighted $d$-Packing}

Algorithm~\ref{alg:DPacking} presents a greedy heuristic for computing a weighted $d$-packing. The algorithm iteratively selects vertices based on a utility function defined as the sum of weights in the closed neighborhood of a vertex. This strategy is inspired by the classical Chvátal’s maximum-utility heuristic for the Set Cover problem, where elements with the largest marginal contribution are selected first.

\RestyleAlgo{ruled}
\SetKwComment{Comment}{/* }{ */}

\begin{algorithm}[hbt!]
\caption{Greedy Weighted $d$-Packing Heuristic\label{alg:DPacking}}
\KwData{
\texttt{$col$ is an array of size $n$ initialized to $0$;} \\
\texttt{queue is an empty queue;} \\
\texttt{$dpack$ is an empty set.}
}
\KwResult{$dpack$}
$x \gets \text{getMaxDegUncol}()$

\While{$x \neq -1$}{
    add $x$ to queue\;
    add $x$ to $dpack$\;
    $col[x] \gets 1$\;
    
    \While{queue is not empty}{
        $v \gets \text{queue.poll()}$\;
        \If{$col[v] \leq d$}{
            \For{each neighbor $w$ of $v$}{
                \If{$col[w] = 0$}{
                    add $w$ to queue\;
                    $col[w] \gets col[v] + 1$\;
                }
            }
        }
    }
    $x \gets \text{getMaxDegUncol}()$\;
}
\Return{$dpack$}
\end{algorithm}

While the pseudocode is self-explanatory, we should note that %The 
array \texttt{col} stores the distance of each vertex from the nearest selected vertex. The function \texttt{getMaxDegUncol} returns an uncolored vertex with maximum neighborhood utility. The algorithm ensures that selected vertices are separated by a distance greater than $d$, while all remaining vertices are dominated within distance $d$.

\paragraph{Computational Complexity.}
Constructing the correlation graph requires $O(p^2)$ time in the number of features $p$, which is unavoidable for pairwise correlation-based approaches. The greedy $d$-packing algorithm runs in $O(|V|+|E|)$ time, as each edge is explored at most once during the breadth-first search procedure. Since subsequent shrinkage methods operate on the reduced feature set, the overall pipeline scales well to high-dimensional data while avoiding the super-polynomial complexity of exact combinatorial formulations.

\subsection{Shrinkage-Based Refinement}

After graph-based pre-selection, we apply shrinkage methods to further refine the feature set. Regularization techniques mitigate overfitting by penalizing model complexity, thereby improving the bias–variance trade-off.

We employ Elastic Net regularization, which combines $\ell_1$ and $\ell_2$ penalties and minimizes
\[
\arg\min_w \left( \sum_{i=1}^n (y_i - f(x_i;w))^2
+ \lambda_1 \lVert w \rVert_1
+ \frac{\lambda_2}{2} \lVert w \rVert_2^2 \right),
\]
where $\lambda_1,\lambda_2 \geq 0$. Elastic Net is particularly effective in the presence of correlated features, as it can retain groups of correlated variables while eliminating redundant ones.

We also consider Support Vector Machine Recursive Feature Elimination (SVM-RFE), which iteratively ranks features according to their contribution to a trained SVM classifier and removes the least informative features.

\subsection{Classification Models}

In the final stage, classification models are trained using the selected features. We use Random Forests and Support Vector Machines with linear kernels, and perform hyperparameter tuning to optimize predictive performance.

\section{Experimental Results}

Our experiments were conducted on six benchmark data sets commonly used in feature selection studies: colon cancer (ColonCA \cite{golub1999molecular} \cite{alon1999broad}), leukemia \cite{golubEsets}, lymphoma, and small blue round cell tumor (SRBCT), obtained from \cite{shu2013}, the Wisconsin Diagnostic Breast Cancer (WDBC) dataset \cite{wdbc}, and a heart disease dataset \cite{heart_disease_45}. These data sets span both binary and multi-class classification settings and are characterized by high dimensionality and strong feature correlations. Table~\ref{tab:summary} summarizes the main characteristics of the gene-expression and WDBC data sets.

For each data set, we computed the coefficient of variation for each feature and used mutual information with the class label as the feature weight. Pairwise Pearson correlation coefficients were then computed between features to build a correlation graph: two vertices are connected whenever the absolute correlation exceeds a given threshold. The resulting weighted graph was processed using our proposed heuristic, henceforth dubbed d-\textsc{WPE}, for varying values of the distance parameter $d$ and the correlation threshold. The selected feature sets were optionally refined using Elastic Net or SVM-RFE, followed by classification. Stability was computed using the same protocol as in \cite{shu2013} to ensure comparability.

\begin{table}[htb!]
\centering
\small
\begin{tabular}{|l|c|c|c|l|}
\hline
Dataset & Features & Instances & Classes & Class Distribution \\
\hline
Colon Cancer & 2001 & 62 & 2 & Normal: 20 -- Tumor: 40 \\
Leukemia & 7128 & 72 & 2 & AML: 25 -- ALL: 47 \\
Lymphoma & 4027 & 66 & 3 & FL: 9 -- CLL: 11 -- DLBCL: 46 \\
SRBCT & 2309 & 83 & 4 & 1: 29 -- 2: 11 -- 3: 18 -- 4: 25 \\
Breast Cancer & 30 & 569 & 2 & Malignant: 212 -- Benign: 357 \\
\hline
\end{tabular}
\caption{Summary of datasets}
\label{tab:summary}
\end{table}

\subsection{Colon Dataset Results}

For the colon cancer dataset \cite{shu2013}, Elastic Net and SVM-RFE without d-\textsc{WPE} reduced the original 2001 features to 11, achieving a classification accuracy of 0.77. Incorporating d-\textsc{WPE} with a correlation threshold of 0.6 and $d=2$ further reduced the feature set to three features, with an accuracy of 0.69. Increasing $d$ resulted in a single selected feature but lowered the accuracy to 0.62.

\begin{figure}[H]
  \centering
  \includegraphics[width=\textwidth]{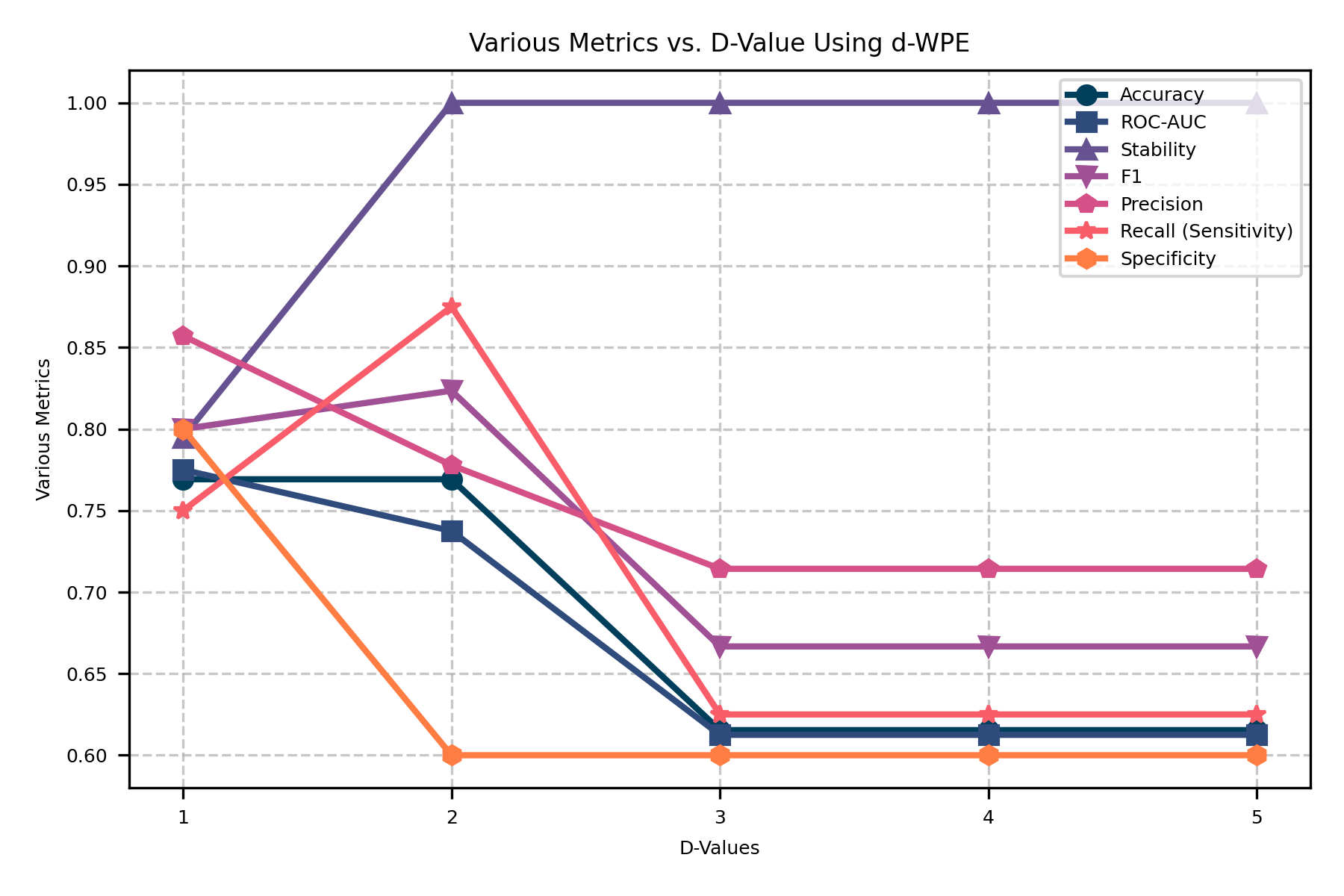}
  \caption{Metric results vs.\ $d$ when applying d-\textsc{WPE} with correlation threshold 0.55 on the colon cancer dataset \cite{shu2013}.}
  \label{fig:colon-plot}
\end{figure}

Reducing the correlation threshold to 0.55 significantly improved performance. At $d=2$, our d-\textsc{WPE} heuristic selected a single feature while maintaining an accuracy of 0.77 and achieving perfect stability. Table~\ref{tab:colon-results} compares these results with competing methods, and Fig.~\ref{fig:colon-plot} shows the evaluation metrics as a function of $d$.

\begin{table}[htb!]
\centering
\small
\resizebox{\textwidth}{!}{
\begin{tabular}{|l|c|c|c|c|c|c|c|c|}
\hline\hline
Method & Corr. Thresh. & Accuracy & Stability & Precision & F1-score & Sensitivity & Specificity & Num. of Features \\
\hline
d-\textsc{WPE} & 0.55 & 0.77 & 1.00 & 0.78 & 0.82 & 0.88 & 0.60 & 1 \\
\hline
d-\textsc{WPE} & 0.60 & 0.62 & 1.00 & 0.71 & 0.67 & 0.63 & 0.60 & 1 \\
\hline
Stability Lasso \cite{shu2013} & 0.60 & 0.83 & 0.61 & -- & -- & -- & -- & 5 \\
\hline
Elastic Net & -- & 0.77 & 0.61 & 0.78 & 0.82 & 0.88 & 0.60 & 23 \\
\hline
Elastic Net with SVM-RFE & -- & 0.77 & 0.72 & 0.78 & 0.82 & 0.88 & 0.60 & 11 \\
\hline
\end{tabular}}
\caption{Comparison of d-\textsc{WPE} vs.\ other methods on the colon cancer dataset \cite{shu2013}.}
\label{tab:colon-results}
\end{table}

\subsection{Leukemia Dataset Results}

\begin{figure}[H]
  \centering
  \includegraphics[width=\textwidth]{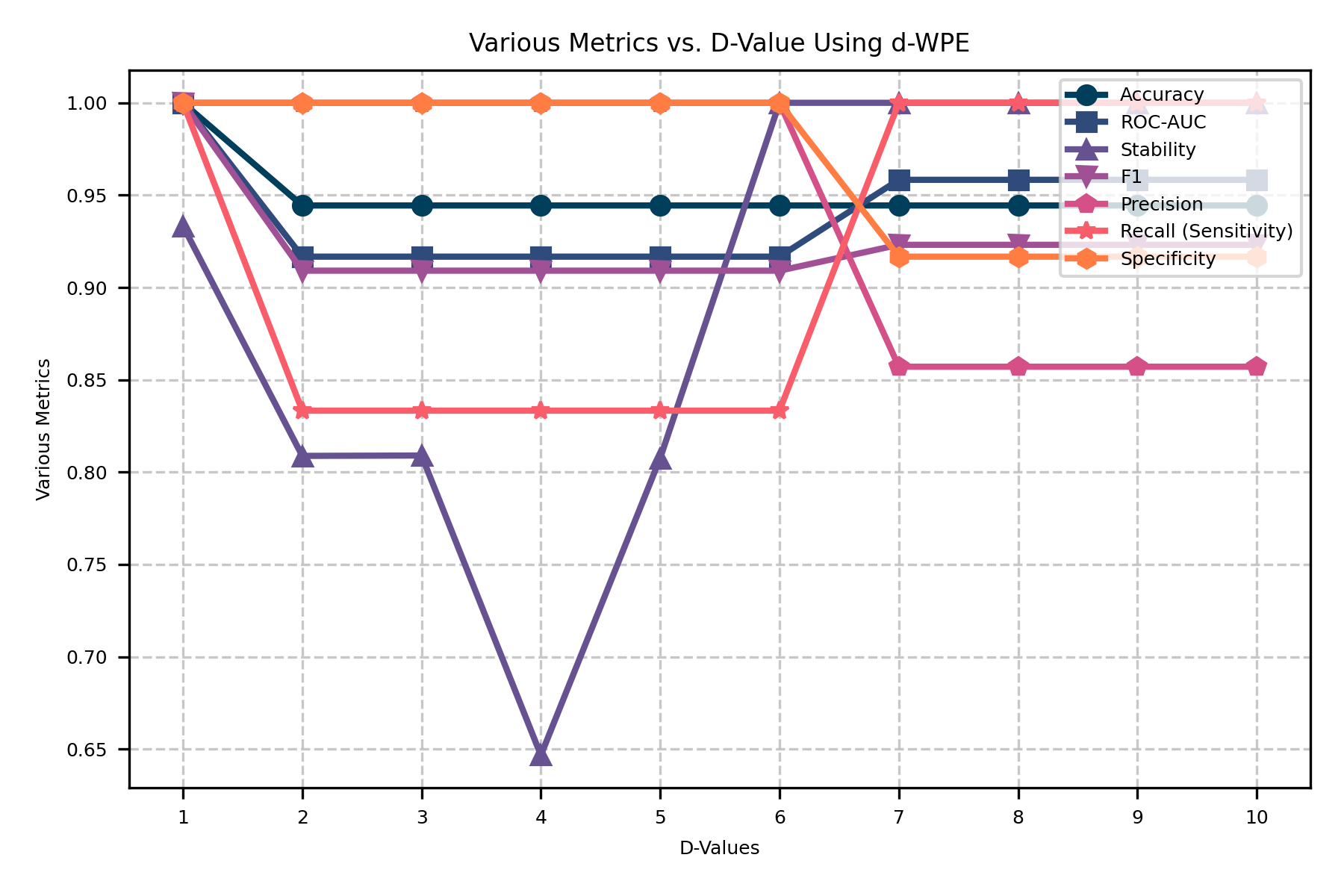}
  \caption{Metric results vs.\ $d$ when applying d-\textsc{WPE} with correlation threshold 0.55 on the leukemia dataset \cite{shu2013}.}
  \label{fig:leukemia-plot}
\end{figure}

Applying d-\textsc{WPE} to the leukemia dataset \cite{shu2013} with a correlation threshold of 0.6 reduced the original 7128 features to a single feature at $d=6$, achieving an accuracy of 0.94 with perfect stability. Lowering the threshold to 0.55 resulted in four selected features with the same accuracy and improved balance across evaluation metrics. Table~\ref{tab:leukemia-results} summarizes the comparison, and Fig.~\ref{fig:leukemia-plot} shows the evolution of the metrics as a function of $d$.

\begin{table}[htb!]
\centering
\small
\resizebox{\textwidth}{!}{
\begin{tabular}{|l|c|c|c|c|c|c|c|c|}
\hline\hline
Method & Corr. Thresh. & Accuracy & Stability & Precision & F1-score & Sensitivity & Specificity & Num. of Features \\
\hline
d-\textsc{WPE} & 0.55 & 0.94 & 1.00 & 0.86 & 0.92 & 1.00 & 0.92 & 4 \\
\hline
d-\textsc{WPE} & 0.60 & 0.94 & 1.00 & 1.00 & 0.91 & 0.83 & 1.00 & 1 \\
\hline
SVM-RFE with MIDS \cite{shu2013} & 0.60 & 0.99 & 0.70 & -- & -- & -- & -- & 14 \\
\hline
Elastic Net & -- & 1.00 & 0.70 & 1.00 & 1.00 & 1.00 & 1.00 & 65 \\
\hline
Elastic Net with SVM-RFE & -- & 1.00 & 0.86 & 1.00 & 1.00 & 1.00 & 1.00 & 32 \\
\hline
\end{tabular}}
\caption{Comparison of d-\textsc{WPE} vs.\ other methods on the leukemia dataset \cite{shu2013}.}
\label{tab:leukemia-results}
\end{table}

\subsection{Lymphoma Dataset Results}

\begin{figure}[H]
  \centering
  \includegraphics[width=\textwidth]{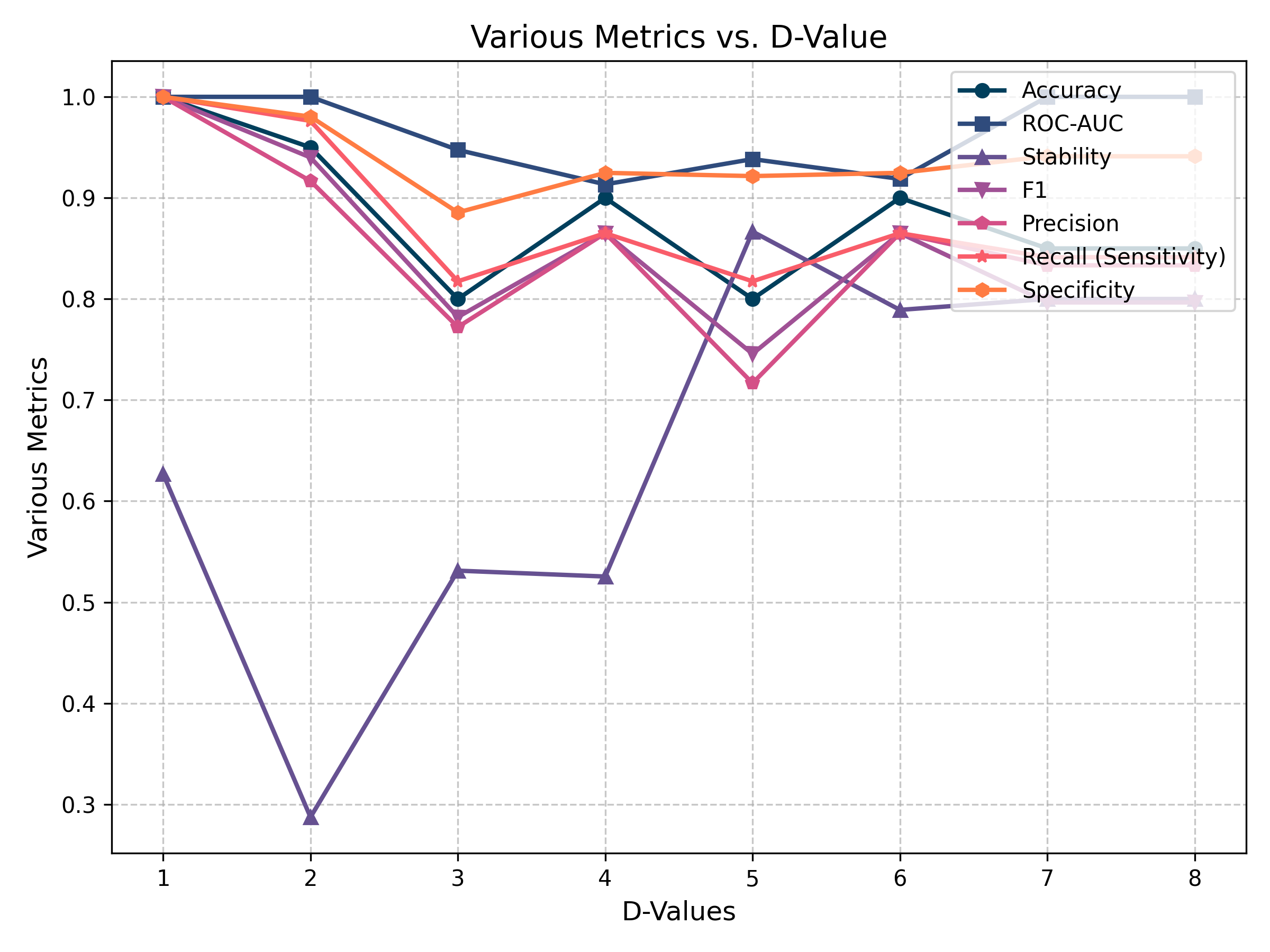}
  \caption{Metric results vs.\ $d$ when applying d-\textsc{WPE} with correlation threshold 0.55 on the lymphoma dataset \cite{shu2013}.}
  \label{fig:lymphoma-plot}
\end{figure}

The lymphoma dataset \cite{shu2013} involves multi-class classification. To accommodate this, Elastic Net and SVM-RFE were applied using a one-vs.-rest strategy. Using d-\textsc{WPE} with a correlation threshold of 0.6 reduced the number of features from 4027 to seven at $d=1$, achieving an accuracy of 0.90. With a lower threshold of 0.55, d-\textsc{WPE} reduced the feature set to three features at $d=5$ while achieving an accuracy of 0.85 and a stability score of 0.80. Table~\ref{tab:lymphoma-results} reports the detailed comparison, and Fig.~\ref{fig:lymphoma-plot} shows the metric behavior with respect to $d$.

\begin{table}[htb!]
\centering
\small
\resizebox{\textwidth}{!}{
\begin{tabular}{|l|c|c|c|c|c|c|c|c|}
\hline\hline
Method & Corr. Thresh. & Accuracy & Stability & Precision & F1-score & Sensitivity & Specificity & Num. of Features \\
\hline
d-\textsc{WPE} & 0.55 & 0.85 & 0.80 & 0.83 & 0.80 & 0.84 & 0.94 & 3 \\
\hline
d-\textsc{WPE} & 0.60 & 0.90 & 0.91 & 0.87 & 0.89 & 0.95 & 0.96 & 7 \\
\hline
SVM-RFE with MIDS \cite{shu2013} & 0.60 & 0.99 & 0.60 & -- & -- & -- & -- & 9 \\
\hline
Elastic Net & -- & 1.00 & 0.47 & 1.00 & 1.00 & 1.00 & 1.00 & 145 \\
\hline
Elastic Net with SVM-RFE & -- & 0.95 & 1.00 & 0.92 & 0.94 & 0.98 & 0.98 & 5 \\
\hline
\end{tabular}}
\caption{Comparison of d-\textsc{WPE} vs.\ other methods on the lymphoma dataset \cite{shu2013}.}
\label{tab:lymphoma-results}
\end{table}

\subsection{SRBCT Dataset Results}

\begin{figure}[H]
  \centering
  \includegraphics[width=\textwidth]{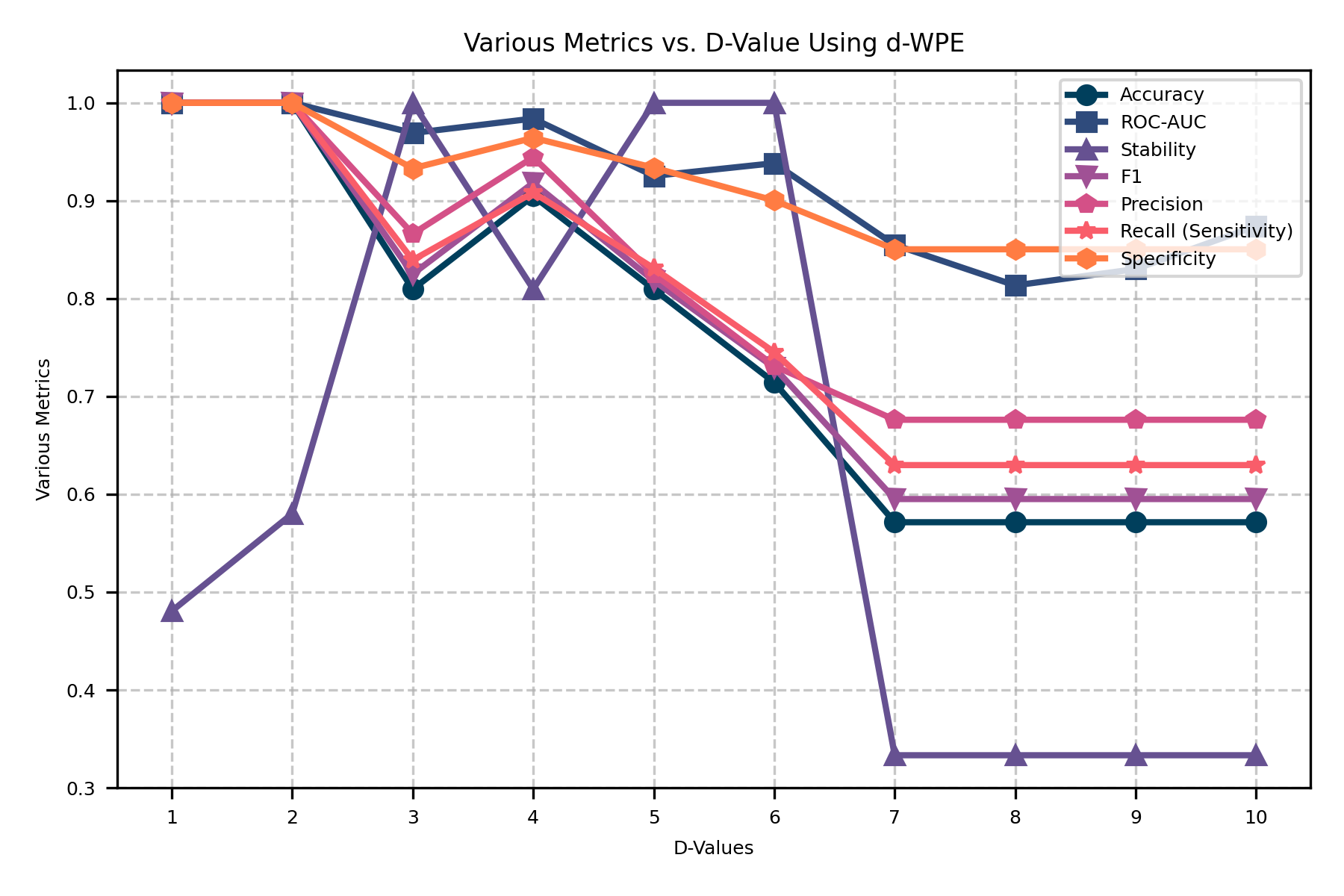}
  \caption{Metric results vs.\ $d$ when applying d-\textsc{WPE} with correlation threshold 0.55 on the SRBCT dataset \cite{shu2013}.}
  \label{fig:srbct-plot}
\end{figure}

We applied the same multi-class procedure to the SRBCT dataset \cite{shu2013}. Using our d-\textsc{WPE} heuristic with a correlation threshold of 0.6, the number of features was reduced from 2308 to 19 at $d=8$, achieving an accuracy of 0.76. With a threshold of 0.55, the method reduced the features to four at $d=5$ with an accuracy of 0.81. Increasing $d$ further reduced the number of features (e.g., to three at $d=7$), but with a substantial drop in accuracy. Table~\ref{tab:srbct-results} and Fig.~\ref{fig:srbct-plot} report the detailed results.

\begin{table}[htb!]
\centering
\small
\resizebox{\textwidth}{!}{
\begin{tabular}{|l|c|c|c|c|c|c|c|c|}
\hline\hline
Method & Corr. Thresh. & Accuracy & Stability & Precision & F1-score & Sensitivity & Specificity & Num. of Features \\
\hline
d-\textsc{WPE} & 0.55 & 0.81 & 1.00 & 0.83 & 0.82 & 0.83 & 0.93 & 4 \\
\hline
d-\textsc{WPE} & 0.60 & 0.76 & 0.78 & 0.80 & 0.77 & 0.78 & 0.91 & 19 \\
\hline
SVM-RFE with MIDS \cite{shu2013} & 0.60 & 0.96 & 0.58 & -- & -- & -- & -- & 24 \\
\hline
Elastic Net & -- & 1.00 & 0.47 & 1.00 & 1.00 & 1.00 & 1.00 & 215 \\
\hline
Elastic Net with SVM-RFE & -- & 1.00 & 0.86 & 1.00 & 1.00 & 1.00 & 1.00 & 14 \\
\hline
\end{tabular}}
\caption{Comparison of d-\textsc{WPE} vs.\ other methods on the SRBCT dataset \cite{shu2013}.}
\label{tab:srbct-results}
\end{table}

\subsection{Breast Cancer Dataset Results}

We tested the proposed method on the WDBC data set \cite{wdbc} using correlation thresholds of 0.55 and 0.70 for graph construction. The results are compared with recent feature selection methods from \cite{WFSIB}, \cite{ABCoDT}, \cite{ELM-RBF}, and \cite{PCA-CA} in Table~\ref{tab:breast_cancer_results}. Using $\tau=0.70$, d-\textsc{WPE} selected three features and achieved an accuracy of 0.9825 with ROC AUC 0.9984. Using $\tau=0.55$, the method selected two features and achieved a perfect ROC AUC, while maintaining competitive performance in the other metrics. We further note that comparison with the previously used methods (in Tables 2-5) was not possible, simply because these methods were never used on the WDBC dataset.

%\todo[inline]{Why the methods in the below table are different from all the previous ones?! Without a proper justification this leads to desk rejection! Justification: The methods here are different because in the previous sections the method we were comparing our findings to was not ran on the wdbc dataset.}

\begin{table}[htb!]
\centering
\small
\resizebox{\textwidth}{!}{
\begin{tabular}{|l|c|c|c|c|c|c|}
\hline\hline
Method & Accuracy & F1-score & Sensitivity & Specificity & ROC AUC & Num.\ of Features \\
\hline
d-\textsc{WPE} (0.55) & 0.9386 & 0.9176 & 0.9070 & 0.9577 & 1.0000 & 2 \\
\hline
d-\textsc{WPE} (0.70) & 0.9825 & 0.9756 & 0.9524 & 0.9825 & 0.9984 & 3 \\
\hline
WFSIB \cite{WFSIB} & 0.9912 & 0.9880 & 0.9858 & 0.9943 & -- & 10 \\
\hline
ABCoDT \cite{ABCoDT} & 0.9718 & -- & -- & -- & -- & 2 \\
\hline
ELM-RBF \cite{ELM-RBF} & 0.9569 & 0.9830 & 0.9756 & 0.9439 & -- & 9 \\
\hline
PCA-CA \cite{PCA-CA} & 0.9900 & 0.9761 & 0.9600 & -- & -- & 10 \\
\hline
Elastic Net & 0.9561 & 0.942529 & 0.953488 & 0.957746 & 0.9971 & 24 \\
\hline
\end{tabular}}
\caption{Comparison of feature selection methods on the breast cancer %(WDBC) 
dataset}
\label{tab:breast_cancer_results}
\end{table}

\subsection{Heart Disease Dataset Results}

We also tested the proposed method on a heart disease dataset \cite{heart_disease_45}. In this case, the method reduced the number of selected features from 13 to 1, and the evaluation metrics remained constant across $d$ values because the computed $d$-packing set did not change. For reference, Noroozi et al.~\cite{Noroozi_Orooji_Erfannia_2023} evaluated several feature selection methods on the same dataset and reported that the smallest feature subset (four features) was obtained via backward selection. In contrast, our method selected a single feature while maintaining comparable performance.

\subsection{Classification Results}

\begin{figure}[H]
  \centering
  \includegraphics[width=\textwidth]{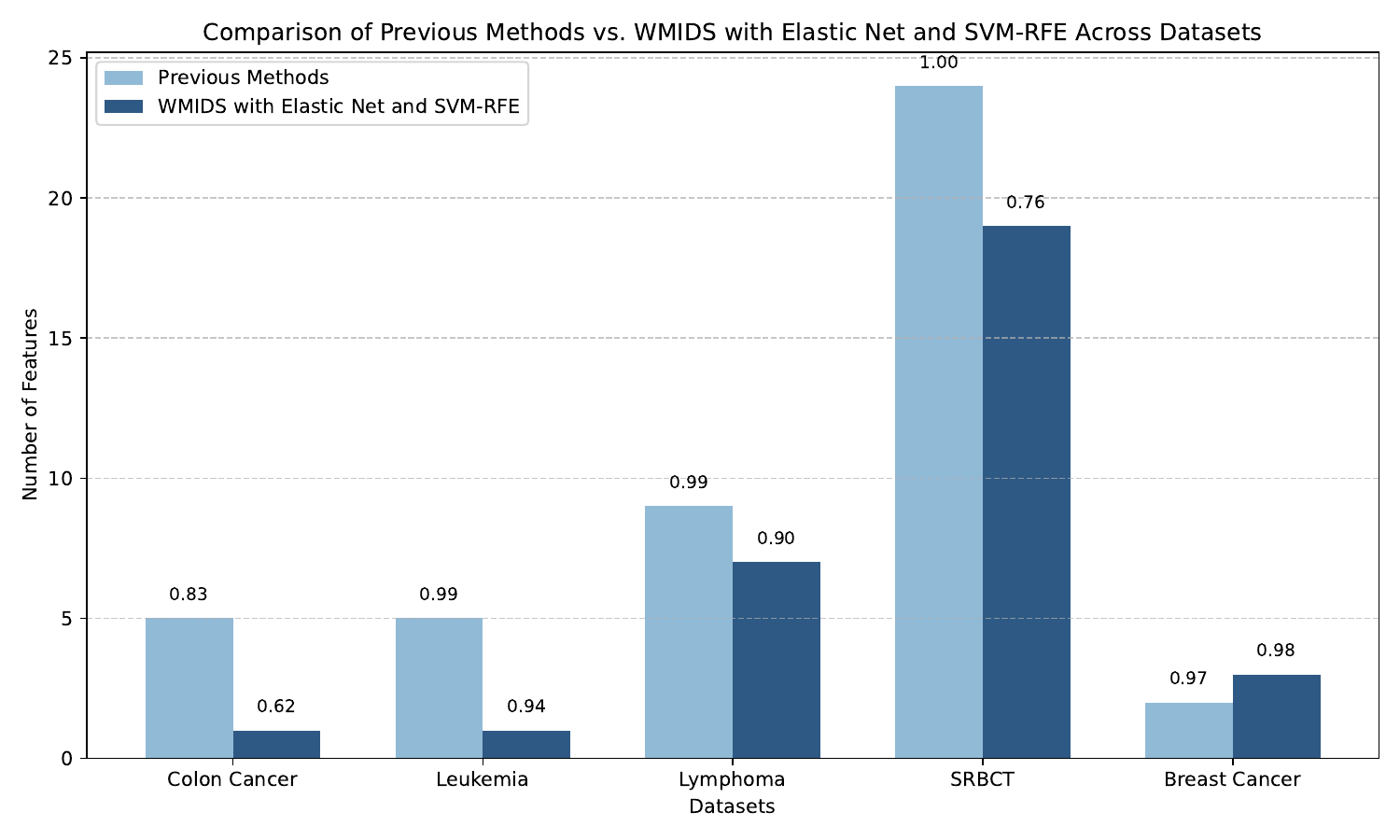}
  \caption{Feature selection comparison with methods from \cite{shu2013} and \cite{10387658} versus d-\textsc{WPE} combined with Elastic Net and SVM-RFE. Classification accuracy is shown above the bars.}
  \label{fig:feature_selection_comparison}
\end{figure}

Overall, d-\textsc{WPE} consistently achieved substantial feature reduction. Figure~\ref{fig:feature_selection_comparison} summarizes the minimum reduced feature counts in comparison with \cite{shu2013} and \cite{abu-wids,10387658}. Our method typically achieves stronger feature reduction with slightly lower accuracy in some cases, but with noticeably higher stability compared to the graph-based approaches in \cite{shu2013}. The difference is particularly pronounced on the leukemia and SRBCT datasets.

The correlation threshold plays a central role in controlling the sparsity (or density) of the correlation graph, and therefore the achievable feature reduction. For example, on the colon dataset, thresholds 0.55 and 0.6 both yield a single selected feature but with different accuracies (Table~\ref{tab:colon-results}). On SRBCT, $\tau=0.60$ yields 19 selected features with accuracy 0.76, whereas $\tau=0.55$ yields four selected features with improved accuracy (Table~\ref{tab:srbct-results}). These observations indicate that $\tau$ should be selected carefully for each data set to balance reduction and predictive performance.

Finally, the choice of vertex weights can also influence results. For instance, on the WDBC dataset, replacing Gini-based weights by mutual information led to a noticeable increase in accuracy, supporting the role of diversity-aware weighting in correlated feature settings.

\subsection{Biological Validation of Selected Features}

To evaluate the biological relevance of features selected by d-\textsc{WPE}, we mapped the selected feature indices back to their corresponding gene identifiers using the original annotation data for each dataset. The colon cancer dataset \cite{alon1999broad} uses UniGene identifiers linked to GenBank accessions, while the leukemia dataset \cite{golub1999molecular} uses Affymetrix HuGeneFL (Hu6800) probe identifiers. Gene annotations were resolved via the Bioconductor hu6800.db annotation package and NCBI Gene linkage.

\paragraph{Colon cancer.}
The two features selected by d-\textsc{WPE} at $\tau = 0.55$ (achieving accuracy 0.70 with a single feature) correspond to:
\begin{itemize}
    \item Feature 266: \textit{CSRP1} (cysteine and glycine rich protein~1, Entrez Gene 1465), a LIM domain protein expressed in smooth muscle cells.
    \item Feature 575: \textit{MORF4L2} (mortality factor 4 like~2, Entrez Gene 9643), a chromatin remodeling factor involved in transcriptional regulation.
\end{itemize}
Notably, \textit{CSRP1} is a smooth muscle marker that lies directly within the primary biological signal identified by Alon et al.\ \cite{alon1999broad}: their analysis found that colon tumor samples were distinguished from normal tissue largely by the reduced expression of smooth muscle genes, reflecting the higher proportion of smooth muscle cells in normal colon tissue. The fact that d-\textsc{WPE} independently selects a gene from this core signal (with no prior biological knowledge encoded in the method) provides strong evidence that the graph-theoretic approach captures genuinely informative features rather than statistical artifacts. A recent study has also reported \textit{CSRP1} as differentially expressed in colon adenocarcinoma \cite{csrp1colon2023}.

\paragraph{Leukemia.}
The features selected by d-\textsc{WPE} at $\tau = 0.60$ (single feature, accuracy 0.94) and $\tau = 0.55$ (four features, accuracy 0.94) map to the following genes:

\begin{table}[h]
\centering
\small
\begin{tabular}{@{}llll@{}}
\toprule
Feature & Gene & Threshold & PubMed hits \\
\midrule
1881 & \textit{CST3} (cystatin C) & $\tau=0.60$ & 6 \\
3344 & \textit{CCR3} (C-C chemokine receptor 3) & $\tau=0.55$ & 8 \\
6166 & \textit{TRDV2} (T cell receptor $\delta$ variable 2) & $\tau=0.55$ & 5 \\
6853 & \textit{SFTPA1} (surfactant protein A1) & $\tau=0.55$ & 0 \\
3548 & \textit{PRAME} (PRAME nuclear receptor) & $\tau=0.55$ & 106 \\
575  & \textit{NUP188} (nucleoporin 188) & plateau & 0 \\
\bottomrule
\end{tabular}
\caption{Leukemia features selected by d-\textsc{WPE}, their corresponding genes, and PubMed citation counts for the gene in the context of leukemia.}
\label{tab:leukemia-genes}
\end{table}

The single feature at $\tau = 0.60$, \textit{CST3}, ranks 49th out of 7129 features in the Golub et al.\ \cite{golub1999molecular} distinction metric, placing it well within the top 1\% of discriminative genes for the ALL/AML classification task. \textit{TRDV2}, a T cell receptor delta chain gene, ranks 82nd. These rankings confirm that d-\textsc{WPE} selects features that are individually highly discriminative.

\section{Discussion of Results}

%The experimental results can be interpreted through the lens of the bias--variance trade-off and, more broadly, statistical learning theory. In graph-based feature selection models such as IDS and $d$-packing, the distance parameter $d$ implicitly controls model capacity by limiting how many features can be selected simultaneously and how correlated they may be. Smaller values of $d$ allow larger feature sets, which may reduce bias but increase variance due to redundancy and correlation. Larger values of $d$ enforce stronger separation constraints, leading to more compact feature sets with reduced variance, potentially at the cost of increased bias.

The experimental results can be interpreted in terms of the bias--variance trade-off and, more broadly, statistical learning theory. In our graph-based feature selection model based on $d$-packing, the distance parameter $d$ implicitly controls model capacity by limiting how many features can be selected simultaneously and how correlated they may be. Smaller values of $d$ allow larger feature sets, which may reduce bias but increase variance due to redundancy and correlation. Larger values of $d$ enforce stronger separation constraints, leading to more compact feature sets with reduced variance, potentially at the cost of increased bias.

Although deriving explicit VC-dimension bounds for classifiers built on $d$-packing–selected features is nontrivial, the empirical behavior observed across all datasets is consistent with this interpretation. In particular, tuning $d$ proved to be an effective mechanism for controlling model complexity in practice. While accuracy sometimes decreased slightly as $d$ increased, stability and specificity consistently improved. These gains are especially important in high-dimensional biological settings, where robustness and reproducibility of selected features are often more valuable than marginal improvements in predictive accuracy.

From the perspective of graph-based modeling, the proposed d-\textsc{WPE framework} generalizes earlier IDS-based approaches by relaxing strict independence constraints. This added flexibility allows the method to better adapt to the structure of correlation graphs derived from real data. Across all evaluated datasets, d-\textsc{WPE} achieved substantial improvements in stability. For example, stability increased by approximately 0.2 on the colon cancer dataset, by about 0.05 on the leukemia dataset, and by roughly 0.6 on the lymphoma dataset for suitable values of $d$. These results suggest that enforcing distance-based separation between selected features is an effective way to mitigate instability caused by highly correlated variables.

The leukemia dataset illustrates particularly well the practical implications of choosing the node packing parameter $d$. In this case, different values of $d$ led to models with distinct trade-offs between accuracy and precision. While one choice of $d$ yielded slightly higher accuracy, another offered better precision and stability. Such situations highlight the importance of combining quantitative evaluation with domain knowledge. Interestingly, the emergence of a single dominant feature at higher values of $d$ may warrant further biological investigation. From a graph-theoretic viewpoint, such a feature could correspond to a structurally central vertex, analogous to those observed in cluster editing models where a single vertex can belong to multiple clusters via the notion of vertex splitting \cite{abukhzam2023cluster}, and may represent a biologically meaningful candidate gene.

From a molecular biology perspective, one of the main advantages of the proposed approach is its ability to drastically reduce the number of selected features. When feature selection models retain hundreds of variables, biological interpretation becomes impractical. By contrast, the compact feature sets produced by d-\textsc{WPE} make downstream analysis feasible. Standard tools can then be applied, including functional annotation and pathway analysis using resources such as NCBI \cite{schoch2020ncbi}, STRING \cite{szklarczyk2023string}, and DAVID \cite{dennis2003david}. When appropriate, these analyses can be complemented by experimental validation, such as survival analysis or Western blot experiments.

%Finally, the results suggest several directions for further improvement. In small-sample, high-dimensional settings, which are common in molecular biology and single-cell data, reducing thousands of features to a few hundred may still be insufficient. One promising direction is the use of minimal $d$-packing \cite{abs-2401-00525}. This imposes even stronger structural constraints. Another is the integration of ensemble techniques such as bagging \cite{gonzalez2020practical} as an intermediate step between graph-based preselection and final regularized feature selection. We are also exploring bipartite graph representations of tabular data, where samples and features form two disjoint vertex sets, enabling feature selection to be formulated as a red--blue dominating set or hitting set problem.

\section{Conclusion}

We introduced a graph-based framework for feature selection based on a weighted $d$-packing model. The key idea is to enforce a controlled separation between selected features while incorporating a notion of diversity through feature-dependent weights.
This combination allows the method to select representative features from correlated groups while favoring those that are more informative with respect to the target variable. The parameter $d$ provides a natural way to control the trade-off between redundancy and coverage, and leads to a more stable selection process in practice.

From an algorithmic perspective, the proposed greedy procedure avoids the need for expensive optimization formulations and scales well to high-dimensional datasets. Experimentally, the method achieves strong feature reduction, improved stability, and competitive predictive performance across several biological benchmarks.

Overall, the results suggest that combining node packing with diversity-aware weighting provides an effective and interpretable approach to feature selection in correlated settings. In contrast to many black-box learning pipelines, the proposed framework offers a transparent selection mechanism in which the role of each feature can be traced back to explicit structural and statistical criteria. This is particularly important in biological applications, where understanding why certain variables are selected is often as valuable as predictive performance itself. More broadly, this work illustrates how simple combinatorial structures can be adapted to capture statistical properties of data in a meaningful and explainable way.

%Overall, the results suggest that combining node packing with diversity-aware weighting provides an effective and interpretable approach to feature selection in correlated settings. More broadly, this work illustrates how simple combinatorial structures can be adapted to capture statistical properties of data in a meaningful way.
%\section{Conclusion}

%This paper presents a graph-theoretic framework for feature selection based on weighted independent dominating sets and their generalization via $d$-packing. Building upon earlier IDS-based approaches, we introduce a diversity-aware weighting scheme and replace computationally expensive integer linear programming formulations with an efficient greedy heuristic.

%Experimental results on multiple high-dimensional biological datasets demonstrate that the proposed d-\textsc{WPE} approach achieves substantial feature reduction while improving stability and maintaining competitive predictive performance. By producing compact, robust, and interpretable feature sets, this framework offers a practical alternative to traditional feature selection methods, particularly in settings characterized by strong correlations and limited sample sizes.

\section*{Acknowledgments}
%This work was supported in part by the Oujiang Lab Start-Up Grant OJQD2022005 funded by the Oujiang lab (Zhejiang lab for regenerative medicine, vision, and brain health), 634 Wenzhou, Zhejiang Province, 325000, China.
This research project was supported by the Lebanese American University under the President’s Intramural Research Fund PIRF0056.

%and minor supported by the Oujiang Lab Start-Up Grant OJQD2022005 funded by the Oujiang lab (Zhejiang lab for regenerative medicine, vision, and brain health), 634 Wenzhou, Zhejiang Province, 325000, China.

%After applying our suite of FPT pattern search tools and the described Bioconductor tools, to help avoid personal bias, generative AI (Claude Opus 4.6) was used in help with the initial interpenetration of the molecular biology results. This interpretation was then enhanced by molecular-biologists (Shaw and).

\end{document}